\documentstyle[aclap]{article}
\title{Probabilistic Event Categorization}
\author{Janyce Wiebe\dag \and Rebecca Bruce\ddag \and Lei Duan\dag \\
\dag Dept. of Computer Science and the Computing Research Laboratory \\
New Mexico State University \\
Las Cruces, NM 88003   \\ 
\ddag Dept. of Computer Science and Engineering  \\
Southern Methodist University \\
Dallas, TX 75275-0112 \\
{\tt wiebe,lduan@cs.nmsu.edu, rbruce@seas.smu.edu} \\
{\it Recent Advances in Natural Language Processing (RANLP-97)},
\\
European Commission, DG XIII, Tzigov Chark, Bulgaria, September 1997, pp. 163--170. }

\begin{document}

\maketitle

\begin{abstract}

This paper describes the automation of a new text categorization task.
The categories assigned in this task are more syntactically,
semantically, and contextually complex than those typically assigned
by fully automatic systems that process unseen test data.  
Our system for assigning these categories uses
a probabilistic classifier, developed with a recent method for
formulating a probabilistic model from a predefined set of potential
features (Bruce 1995, Bruce and Wiebe 1994, Pedersen et al. 1996).
This paper focuses on feature selection.  It presents various types
of properties experimented with in this work.
We identify and evaluate various approaches to organizing
the collocational properties into features.  With the more complex
features we define, there is an organization that yields the best
results; but the same organization with less complex features yields
inferior results.  The results suggest a way to take advantage of
properties that are low frequency but strongly indicative of a class.
The problems of recognizing and organizing
the various kinds of
contextual information required to perform a linguistically
complex categorization task has rarely been systematically
investigated in NLP.  
\end{abstract}
\section{Introduction}
\label{intro}
This paper reports findings on performing automatic event
categorization, i.e., recognizing high-level semantic classes of the
main state or event that a clause is about.  The event categorization
addressed in this paper is new.  In this classification scheme, the
event reported by the main clause of a sentence is categorized as
being either: (1) a private state (the clause is about, e.g., a
belief, emotion, or perception), (2) a speech event (the clause is
about, e.g., a saying or declaring event), or (3) other (the clause is
about another kind of state or event).  The speech-event category is
divided into subcategories based on how the event is presented
syntactically and how much of what was said is presented in the
sentence.
The language used to describe private states and speech events is rich
and varied, including idiomatic and metaphorical expressions (Barnden
1992).  There is a large amount of syntactic and part-of-speech
variation, and the categorization is context dependent.
Although the categories are
complex, it has been demonstrated in an inter-coder reliability study
(Wiebe and Bruce 1997) that these classifications can be performed with
high reliability by human judges.

The method we use to automate this task is probabilistic
classification.  We perform an explicit model search to find a model
that provides a good characterization of the relationships among the
targeted classification and properties in the data (Bruce 1995, Bruce
and Wiebe 1994, Pedersen et al. 1997).  Doing so is in contrast to one
common practice in NLP of assuming a certain model form, such as
n-gram and Naive Bayesian models, without testing how well those
models fit the data.  In the experiments reported on here, the models
identified as best for the task being performed vary in structure in
response to the type of features used, supporting the usefulness 
of performing model search.  
The method permits the use
of many features of different kinds, including n-gram properties
as well as those types of features 
typically included in Maximum Entropy models (Berger et al. 1996)
and Decision Trees (Breiman et al. 1994).
In addition, as in Decision Tree
induction (Breiman et al. 1994), 
feature selection can be performed as part of
the process of model formulation.

We experimented with many different kinds of properties to perform the
classification task.  These properties are presented in this paper.
They are determined fully automatically, and 
range 
from shallow surface characteristics (e.g., word counts
and word co-occurrence) to more syntactically complex structures (e.g.,
an adjective serving as subject complement) as well as discourse
features (e.g., whether or not the sentence is the first one in a
paragraph). 
Many of the properties would be applicable to other event
categorization and information extraction tasks for which one event
out of many in a sentence is targeted, or for which the
classifications are highly context dependent.

We also experimented with various ways to organize collocational
properties into features, including properties that are
often used in word-sense disambiguation systems.
With the more complex properties we define,
there is an organization that yields the best results, but with the
less complex properties, the same organization yields inferior results.
The results suggest a way to take better advantage of properties that
are low frequency but strongly indicative of a class.    

In
addition to such factors as the form of the model and the method used
to choose collocations, the organization used for collocational
information is another experimental parameter that can affect
performance of an NLP system that uses collocational
information.

A preprocessor was developed to determine the properties according to
which the classifications are made.  It is composed of 
off-the-shelf components
and new components.  
The new components and pointers
to the existing ones 
will be available over the World Wide Web.  The
annotation instructions, the results of the intercoder-reliability
study,  and tables of feature
values for experimentation will also be available at that
site.

The remainder of this paper is organized as follows. 
The method used for model selection is described in section
\ref{method}. The results of the experiments are given up front in
section \ref{results}, and then discussed in subsequent sections.
Section \ref{features} details the properties experimented with, and
section \ref{organ} presents different possible organizations of contextual
information into features. Section \ref{discussion} discusses the
results, and section \ref{futurework} is the
conclusion.
\section{The Method}
\label{method}
We use a supervised learning method for automatically formulating
probabilistic models for use in classification, where a classifier is
induced from a corpus of tagged data.
Suppose there is a training sample in
which each sentence is represented by the variables $(F_1,
\ldots, F_{n-1}, S)$.  Variables $(F_1,\ldots, F_{n-1})$ correspond to
properties of the sentence and the context in which it appears, and
variable $S$ is the classification variable, the variable that
corresponds to the classification being made.  Our task is to induce a
classifier that will assign a value for $S$, given the values that the
feature variables have for this sentence.

We adopt a statistical approach whereby a probabilistic model is
selected that describes the interactions among the feature variables.
This approach is described fully elsewhere (Bruce 1995, Bruce and Wiebe
1994, Pedersen et al. 1996).  Such a model can form the basis of a
probabilistic classifier since it specifies the probability of
observing any and all combinations of the values of the feature
variables.

In the fully saturated model, all variables are interdependent, and
the parameters of the model correspond to combinations of values of
all of the variables in the model.  If the data sample can be
adequately characterized by a less complex model, i.e., a model in
which there are fewer interactions between variables, then more
reliable parameter estimates can be obtained.  How well a model
characterizes the training sample is determined by measuring the {\it
fit} of the model to the sample, i.e., how well the distribution
defined by the model matches the distribution observed in the training
sample.

A good strategy for developing probabilistic classifiers is to perform
an explicit model search to select the model to use in classification.
The model selection algorithm 
used here performs a backward sequential search (a type of greedy
search) of the class of {\em decomposable models}, a class of models
that have many computational advantages (Whittaker 1990).

A backward sequential search is performed, which 
begins by designating the saturated
model as the current model.  At each stage, we generate the set of
decomposable models of complexity level $i-1$ that can be created by
removing an edge from the current model of complexity level $i$.  The
evaluation criterion is applied to each of these models to determine
which yields the least degradation in fit from the current model.  If
the degradation is within limits established by the evaluation
criterion, this becomes the current model and the search continues.
Otherwise, the search stops.
For a further discussion of search strategies and evaluation criteria,
see Pedersen et. al. (1997).

The model selection process also performs feature selection. If a
model is selected where there is no edge connecting a feature variable
to the classification variable, then that feature has been, in essence,
dropped from the classifier.  
The Log-likelihood ratio statistic $G^2$
(Bishop et al. 1975) is used as the model evaluation criterion in all
of the experiments.  

\section{The Experiments and Results}
\label{results}

This section presents the results of the comparative experiments
performed in this paper.  The properties used to form features are
presented in section \ref{features}, and the various organizations of
collocational information are given and discussed in section \ref{organ}.

After a large amount of background experimentation, the best
experiment we found involves:
(1) four non-collocational features (those labeled the {\it Current
Best} in section \ref{features}), and (2) the collocational properties
labeled {\it Syntactic Patterns} in section \ref{features},
organized as {\it per-class-2}, which is described
in section \ref{organ}.
A feature was judged to be good if, after the model search procedure
has completed, that feature is still included in (one of) the model(s)
with the highest accuracy.

Since our interests are to investigate the relative goodness of the
various collocational patterns and of the organizations, we varied
only these factors, and used the same set of non-collocational
features throughout.

The total amount of data 
consists of 2,544 main clauses from the Wall Street Journal
Treebank corpus (Marcus et al. 1993).
The distribution of
classes over the entire data set is shown in table 1.
The lower bound for the problem--- 
the frequency in the entire data set of the most frequent class (Gale
et al. 1992b)---is 52\%. 

\begin{table*}
\begin{center}
\caption{Distribution of Classes}
\vspace*{6mm}

\begin{tabular}{|l|c|}
\hline
\multicolumn{1}{|c|}{Class} &
\multicolumn{1}{c|}{Percentage of the Corpus}  \\
\hline
{\it Private state} &  10\%\\
 & \\
\hline
{\it Speech category 1:} &   09\% \\
\hspace*{4mm} {\it direct speech} & \\
\hline
{\it Speech category 2:} & 04\% \\
\hspace*{4mm} {\it mixed direct and indirect speech} & \\
\hline
{\it Speech category 3:}  & 24\% \\
\hspace*{4mm} {\it other speech event}  & \\
\hline
{\it Borderline private state and other event} & 01\%\\
  & \\
\hline
{\it Other state or event} & 52\%\\
  & \\
\hline
\end{tabular}
\end{center}
\end{table*}

For clearer understanding of the factors covaried in
the experiments presented in table 2, 
the model search procedure was not permitted
to drop any features from the model.

10-fold cross-validation was performed.
For each fold, the collocations were determined and
model search was performed anew.
Each fold
is a different split between 1/10th testing data ($Test Data$), 
and 9/10th training
data ($Training Data$).  
For each fold, $Training Data$ was further split into 9/10th 
training
data ($Search Data$; 81\% of the total data) and 1/10th test data 
($Selection Data$; 9\% of
the total data).  Model search was performed on $Search Data$,
and the model $M$ with 
the highest accuracy on $Selection Data$ was selected.
Finally, the accuracy, precision, and recall of Model $M$
on the real test set, $Test Data$, were determined; 
the results
presented in table 2 are the averages of 
those results over all of
the folds.   Thus, which model to choose as best is
based on a search-selection split of the training
data, and the results are reported on separate, held-out test data.

In table 2, rows correspond to the organizations defined in section
\ref{organ}: (PC-1 for {\it per-class-1}; PC-2 for {\it per-class-2};
OR-1 for {\it over-range-1}, and OR-2 for {\it over-range-2}). Columns
correspond to collocation 
types.

\begin{table*}

\caption{10-fold Results Varying Collocation Type and Feature
Organization }

\vspace*{5mm}
\begin{tabular}{|cccc|ccc|ccc|}
\hline
& & & & & & & & & \\
& \multicolumn{3}{c|}{Co-occurrence Patterns} & 
  \multicolumn{3}{c|}{Within-5 Patterns} & 
  \multicolumn{3}{c|}{Syntactic Patterns} \\
& & & & & & & & & \\
\hline
& & & & & & & & & \\
& Accuracy & Precision & Recall & 
Accuracy & Precision & Recall  &
Accuracy & 
Precision & Recall \\
& & & & & & & & & \\
\hline
& & & & & & & & & \\
OR-1  
& 0.6838 & 0.6967 & 0.9815 
& 0.6020 & 0.6144 & 0.9799 
& 0.7039 & 0.7056 & 0.9976 \\
& & & & & & & & & \\
\hline
& & & & & & & & & \\
OR-2  
& 0.7063 & 0.7164 & 0.9858  
& 0.7082 & 0.7147 & 0.9909 
& 0.7114 & 0.7158 & 0.9937  \\
& & & & & & & & & \\
\hline
& & & & & & & & & \\
PC-1 
& 0.5315 & 0.5364 & 0.9906 
& 0.5550 & 0.5568 & 0.9969 
& 0.7382 & 0.7431 & 0.9933  \\
& & & & & & & & & \\
\hline
& & & & & & & & & \\
PC-2 
& 0.6500 & 0.6571 & 0.9886 
& 0.6567 & 0.6604 & 0.9945 
& 0.7468 & 0.7495 & 0.9965         \\
& & & & & & & & & \\
\hline 
\end{tabular}
\end{table*}

A better result than any in the table was obtained in a separate
experiment, in which some hand-tuning of the collocational features was
performed: over 78\% by manually grouping some related information
into features.

\section{Properties}
\label{features}
The properties we experimented with are given in this section,
along with brief indications of the preprocessing required to
determine them.  Many are similar to the kinds of surface properties
suggested by Hearst (1992) and Light (1996).  Some are based on
properties found to be correlated with similar classes in the
literature; others are based on observing the tagged training data;
and others were chosen based on intuition and the fact that the
preprocessor is able to determine them (such as the tense of the main
verb).

The Treebank syntax trees were used for only one purpose, to identify
the main clause of the sentence.  The reason that the main clause must
be identified is only because we define the problem as classifying the
main clause.  
The features could easily be adapted to any clause,
whether or not it is the main clause.

The main verb of the clause to be classified is the pivot of some of
the properties.  We adopt Quirk et al.'s definition of a main verb
(1985), and use a finite-state machine to skip over the
various types of auxiliaries and identify the main verb automatically.
In identifying and applying the collocational properties listed below,
the morphological analyzer described in Karp et al. (1992) is used to
match the root forms of words, and 
Brill's tagger (Brill 1992) is used to
assign parts of speech.

We begin with the non-collocational properties, listing first those
from the best experiment we found.  Listed second are properties that
were chosen in some experiment for inclusion in the most accurate
model.  This occurred either on the current data with a different
subset of features than those in the best experiment, or on an earlier
version of the annotated data.  In this earlier version of the problem
definition, the annotations were less context-sensitive, and the task
was more like traditional word-sense disambiguation.  Listed third
are those we did not succeed with.

\subsection{Non-Collocational Properties}
\subsubsection{The Current Best}
The following non-collocational features are the best we found for the
current problem.
\begin{enumerate}
\item
Whether or not the sentence begins a new paragraph.  Paragraphs are
already delimited in the Treebank corpus.

Psychological experiments have shown a correlation between paragraph
breaks and point of view sentences (Stark 1987, Bruder and Wiebe 1990).
That this is one of the best features lends further
support to those findings.
\item
Percentage of the sentences so far in the current paragraph that the
system classified as private-state or speech-event sentences.
The value is 1 if this
proportion is greater than 0.3, 0 otherwise.
The goodness of this feature also gives evidence for the importance of
the paragraph as a unit for this problem.  
\item
Define a quote ratio,  $R=N/M$, where $N$ is the number of words
which are within quotation marks in the sentence, and $M$ is the
total number of words in the sentence.  
There are three levels to this property:
$R$ greater than 0.3;  $R$ between 0.3 and 0.1; and
$R$ less than 0.1.

\item
Whether or not ``according to" appears.  
\end{enumerate}

{\bf Good in other experiments}
\begin{enumerate}
\item
WordNet synsets (Miller 1990).  This property was motivated by uses of
WordNet synsets in Resnik (1993) and Roget categories in Yarowsky
(1992).

For abstract classes we need to extend coverage beyond individual word
collocations.  Thus, we experimented with the following synset
properties.  Let $W$ be a set of words chosen as collocations
in some manner (see sections \ref{colprops} and \ref{organ}).
A synset
property is whether or not there is a member of the same synset as a
member of $W$ in the sentence (keeping to the same part of speech).
\item
The class assigned by the system to the previous sentence, i.e.,
a 2-gram property.
\item
Whether or not the subject of the main clause contains 
a proper noun.
\item
Whether or not the subject of the main clause contains
a personal pronoun.

The preprocessor uses the output of a proper name recognizer developed
by Jim Cowie at the Computing Research Laboratory at NMSU. 
\item
A set of binary properties, each mapped to its own feature:
``that'' appearing within a window after the main verb of the
main clause; a comma appearing before the main verb;
and a colon appearing just after the verb.   We intend, in the near
future, to treat these the same way that collocations are
treated (see section \ref{colprops} on collocational properties).
\item
The tense of the main verb.
\item
The absence or presence of  
``to'' followed by the pattern
NPapprox-short within X words (+ or -) of the main verb, where 

NPapprox-short = det* adj* $\mbox{noun}^+$ adj* 

Example: ``The company looked attractive to the investors''. 
\end{enumerate}

\subsubsection{Not found to be useful}
\begin{enumerate}
\item 
The length of the current sentence (above or below a threshold). 

This property was meant to be an approximation of whether or not the
sentence is a complex sentence.
\item
The number of sentences in the current article (above or below a
threshold).

This is a property of the entire article.  The intuition is that
longer articles are more likely to express reactions to events and
motivations for actions.  The difficulty of such properties for
supervised learning methods is data sparsity, since the objects are
entire articles rather than sentences.

\item
The total number of proper nouns in the article,
another property of articles rather than sentences.
\end{enumerate}

\subsection{Collocational Properties}
\label{colprops}
By {\it collocation} we mean a relationship between a word and the
annotation class.  
In this study, we consider a range of collocational patterns, from
simple co-occurrence to those defined by syntactic
expressions.  Like many others (e.g., Hearst 1992, Berger et al. 1996,
Robin 1996, Golding and Schabes 1996), our best results were obtained
using collocations based on regular expressions composed of
part-of-speech 
tags
and the root forms of words.  Such collocations can better pinpoint a
particular state or event out of all those referred to in the
sentence.
When one event is being targeted, as
in information extraction and event categorization, there is often
noise if the entire sentence is considered.

Our syntactic collocational patterns are defined in section 
\ref{synpats} below.  
These patterns define basic syntactic structures that are not specific
to our particular problem.  

In addition to the syntactic patterns, we also
experimented with two simpler collocational patterns that are
commonly used in NLP.  These are presented in sections \ref{within-5}
and \ref{co-occurrence}.

Below, the symbol {\it main\_verb-MC} refers to
the main verb of the main clause, and
{\it NPapprox} is defined as follows:
NPapprox = NPapprox-short $\mid$ NPapprox-short prep NPapprox. \\

\subsubsection{Syntactic Patterns.}  
\label{synpats}

\vspace*{3mm}
\noindent
baseMVColPat = $\{$v $\mid$ v is main\_verb-MC$\}$.  \\
E.g., ``She believes that Mary is sweet.''  \\

\noindent
baseAdjColPat = $\{$a $\mid$ a is in the 
pattern $\langle$ main\_verb-MC adv* a$\rangle$, 
where the main verb is copular$\}$ \\
E.g., ``She is/seems happy'' \\

\noindent
complexMVColPat = $\{$v $\mid$ v is in the pattern 
$\langle$main\_verb-MC adv* [ NPapprox ] [ ``to'' ]
v$\rangle$,  where v is a main verb$\}$ \\
E.g., ``He made her jump.'' \\

\noindent
complexAdjColPat = 
$\{$a $\mid$ a is in the pattern 
$\langle$main\_verb-MC adv* [ NPapprox ] [ ``to'' ]
adv* v adv* a$\rangle$, where
v is a main\_verb and v is copular$\}$ \\
E.g., ``He tried to be happy'' or ``It lead him to possibly be very
happy.'' \\

We also experimented with noun syntactic patterns, but 
did not identify any that improved performance.

\subsubsection{Within-5 Patterns.} 
\label{within-5}

\noindent
One for each of verbs, nouns, and 
adjectives: \\
Within-5 = $\{$w $\mid$ w appears within 5 words (+ or -) of 
main\_verb-MC$\}$.

\subsubsection{Co-occurrence Patterns.}  
\label{co-occurrence}

\noindent
One for each of 
verbs, nouns, and adjectives: \\
Co-occurrence = $\{$w $\mid$ w appears anywhere in the sentence$\}$.

\section{Selecting Collocations and Organizing
Information into Features}
\label{organ}
There are a number of ways to organize collocational properties, such
as those defined above, into features.  To produce the results
presented above in section \ref{results}, we systematically varied the
type of organization used.

The patterns defined above are used in combination with a selection
method to identify individual collocations.  The organization of the
collocations into features and the method used to identify the
individual collocations are interdependent.  Let there be $c$
annotation classes, $C_1$ to $C_c$.  Let there be $p$ collocational
patterns, $P_1$ to $P_p$ (e.g., {\it baseMVColPat} 
is one such pattern).  

Then there are two ways to select collocations: (1) select words that
are correlated with class $C_i$ when they appear in pattern $P_j$;
these are referred to as {\it per-class collocations}, and are denoted as
$WordsC_{i}P_{j}$; and (2) select words that, when they appear in
pattern $P_j$, are correlated with the classification variable
across its entire range of values.  These are referred to as {\it over-range
collocations}, and are denoted as $WordsP_{j}$.

\subsection{Identification of Per-Class Collocations}
\subsubsection{Criterion for Identifying Collocations.}
The method 
used here and in Ng and Lee (1996) for forming the
collocation sets $WordsC_{i}P_{j}$ is (in the experiments, we use k =
0.5):

\[ WordsC_{i}P_{j} = \{w \mid P(C_{i}|w \mbox{ in } P_j) > k \} \]

\subsubsection{Organizations}

We experimented with two organizations that are in greatest contrast
with the over-range organizations given below.

\noindent
{\bf Organization per-class-1}
There is one binary feature for each class $C_i$, whose value
is 1 if any member of any of the sets $WordsC_{i}P_{j}$ 
appears in the sentence, $1 \leq j \leq p$. 

\noindent
{\bf Organization per-class-2}
For each pattern $P_j$, define a feature with $c+1$ values as follows:\\
For $1 \leq i \leq c$, there is one value which corresponds to 
the presence of a word in $WordsC_{i}P_{j}$.  Each feature also has a value
for the absence of any of those words. \\

\subsection{Identification of Over-Range Collocations}
\subsubsection{Criterion for Identifying Collocations.}
In this alternative, the members of the collocation sets $WordsP_{j}$
are identified as follows.  $G^2$ (or another goodness-of-fit test) is
applied to identify words $w$ such that, when $w$ appears in pattern
$P_{j}$, the model of independence between the classification variable
and $w$ has a poor fit.

\noindent
{\bf Organization over-range-1} \\ This organization is used in
positional features such as in Gale et al. (1992a) and Leacock et
al. (1993).  Define one feature per pattern $P_{j}$, with $\mid
WordsP_{j} \mid + 1$ values, one value for each word in $WordsP_{j}$
(i.e., each word selected for pattern $P_{j}$ using $G^2$ as described
above).  Each feature also has a value for the absence of any word in
$WordsP_{j}$. \\

\noindent
{\bf Organization over-range-2} \\ This organization is commonly
used in
NLP.  Define a binary feature for each word in each set $WordsP_{j}$, \\
$1 \leq j \leq p$.

\section{Discussion}
\label{discussion}
As can be seen in table 2, the best results are obtained with
the {\it per-class-2} organization, which is not commonly used in NLP.

Notice in table 2 that good results are obtained with the per-class
organizations and the syntactic patterns.  But poorer results are
obtained with the per-class organizations and the simpler
collocational patterns.  The simpler collocational patterns can give
relatively good results---they do so when used with the over-range
organizations.

\vspace*{3mm}
\begin{center}
Table 3: Positive and False Positive Occurrences of \\ 
Collocational Features using Organization PC-1 \\
(averages across features)

\vspace*{5mm}
\begin{tabular}{|lc|c|}
\hline
 & Total Positive &  False Positive  \\
N=255 &  Instances     &     Instances \\
\hline
Co-occurrence   &         84                      &     52  \\
Within-5  &               73  &             41  \\
Syntactic patterns &    21           &       7 	 \\
\hline 
\end{tabular}
\end{center}

\vspace*{5mm} 

In comparison to the more restrictive (syntactic) patterns, the
less restrictive (co-occurrence and within-5) 
patterns identify properties that occur more
frequently, but do not as strongly select one of the classes.  To see
this, consider table 3, which contains frequency information
for one of the folds of the experiments whose results
are in table 2, row 3. 
The first column shows that 
the total number of positive instances is much higher for the less
restrictive collocational patterns than for the more restrictive ones.
The second column shows that the number of false positives (e.g.,
a ps collocation that appears with a class other than ps) is also
much higher for the less restrictive collocational patterns.

Organization {\it per-class-1} admits the least amount of interaction
between the words in the collocation sets and the other features: all
the collocation words are grouped into one value of one feature.  The
less restrictive properties benefit from the organizations that permit
more interaction.  In interaction with other features, these
properties becomes stronger indicators of a specific class.

With the over-range organizations, the syntactic patterns lead to many
variable values for which there are seldom positive instances (since
even grouped together, the frequency is low, as table 3 shows).  The
experiments presented in table 2 demonstrate that having many 
variables that contribute no evidence for most instances can harm
accuracy.  Methods have been proposed for handling low-frequency,
highly indicative properties.  One is to consider only collocations
that occur above some threshold frequency (e.g., Smadja 1993 and Ng
and Lee 1996).  However, it is desirable to be able to retain these
words, because when they occur, they are good indicators.  
Hearst (1992) addresses this problem by considering only positive
evidence.  Similarly, Yarowsky (1993) considers only the single best
piece of evidence that occurs.  Another way to handle this problem is
the one presented here: by identifying the collocations using the
per-class method, one is able to retain low-frequency,
highly indicative properties by consolidating them into fewer
variables.
\section{Conclusion}
\label{futurework}
This paper presented the results of a study in which a fully automatic
system for event categorization was developed and tested.  The system
was developed using a recent method for formulating a probabilistic
model to use in classification.  Although the categorization task is
complex, 10-fold cross validation results were presented, showing good
performance: 75\% accuracy, which is a 44\% improvement over the lower
bound.  Some manual tuning of features raise the results above 78\%.

Our focus in this paper was feature selection.  
Many different contextual properties
were described and evaluated.  The features evaluated in this study
would be applicable to other event categorization and
information extraction tasks for which one event out of many in a
sentence is targeted, or for which the classifications are highly
context dependent.  In future work, we plan to investigate
including the additional features that Siegel (1997) and Klavans \& Chodorow 
(1992) found to be important for state versus event classification.

In addition to identifying relevant contextual properties, contrasting
approaches to organizing collocational properties into features were
defined and systematically tested.  The results suggest that a
grouping of features allowing fewer interactions is desirable for low
frequency, highly indicative properties.
On the other hand, the results suggest
that higher-frequency, less indicative properties yield better
results when the information is organized so that a greater degree of
interaction among variables can be exploited.  While these findings
were obtained using a particular method for model selection, they
should be equally applicable to any classification system that allows
interactions among features and supports the types of features described
in this study.  
\section{Acknowledgements}
This research was supported in part by the Office of Naval
Research under grant number N00014-95-1-0776.
We thank Julie Maples for her work developing the annotation
instructions and manually annotating the data, and Kenneth
McKeever and Tom O'Hara for helpful insights regarding 
evaluation.

\end{document}